\documentclass{article}
\usepackage{times}
\usepackage{helvet}
\usepackage{courier}
\usepackage{xcolor}
\usepackage{float}
\usepackage{amssymb}
\usepackage{amsmath}
\usepackage{geometry}
\usepackage{graphicx}
\graphicspath{./img/}

\usepackage{biblatex}
\nocite{*}

\begin{document}

\title{C-Causal Blindness}
\author{Gonçalo Hora de Carvalho, \\goncalo@iiim.is}
\maketitle

\begin{abstract}
\begin{quote}
This text is concerned with a hypothetical flavour of cognitive blindness referred to in this paper as \textit{C-Causal Blindness} or C-CB. A cognitive blindness where the policy to obtain the objective leads to the state to be avoided. A literal example of C-CB would be \textit{Kurt Gödel's} decision to starve for \textit{"fear of being poisoned"} - take this to be premise \textbf{A}. The objective being \textit{"to avoid being poisoned (so as to not die)"}: \textbf{C}, the plan or policy being \textit{"don't eat"}: \textbf{B}, and the actual outcome having been \textit{"dying"}: $\lnot$\textbf{C} - the state that Gödel wanted to avoid to begin with. Gödel pursued a strategy that caused the result he wanted to avoid. An experimental computational framework is proposed to show the isomorphic relationship between C-CB in brain computations, logic, and computer computations using a new proposed algorithm: a Weighted Hidden Markov Model.
\\
\end{quote}
\end{abstract}

\section{Introduction}
This text is concerned with exploring a hypothetical cognitive bias: \textit{C-Causal Blindness} (C-CB). Its nature is hypothesised to be a bundle composed of some subset of cognitive fallacies known to exist in some form or another in decision-making \cite{heuristicsbiases}. C-CB, a flawed thinking pattern or bias, has found its way into multiple adages throughout history - "universal truths" that people have packaged in the form of sayings that often relate to \textit{self-fulfilling prophecies}. In modern times, it has been put succinctly by the French fabulist Jean de la Fontaine (1621--1695) and summarised in the children's movie \textit{Kung Fu Panda}. It was reiterated very eloquently by the character of Master Oogway, an old wise turtle who happens to be a monk, in the form of a proverb: "One often meets his destiny on the road he takes to avoid it...". As a cognitive bias bundle, C-CB encapsulates the types of cognitive processes that are concerned with producing actions quickly while favouring simple or elementary decision nodes from the great tree of possible actions. It probably involves fallacies or mental processing mistakes where one supports an idea or regards it as better than others simply because of its novelty (commonly referred to as the \textit{appeal to novelty} fallacy). It might also be related to the \textit{ambiguity effect}, which is regarded as a cognitive bias that, as a heuristic, functions to avoid decision-making branches that lack information or for which the probability of a positive outcome is unknown. 

Recent coalitions of computational and theoretical frameworks have yielded great predictive results in explaining thought mechanics. One such framework is the Bayesian brain, where erroneous heuristics and biases, including availability, representativeness, and anchoring, are explained by modelling the brain sampling from memory or the creation of mental representations which diverge from formal probabilistic calculations \cite{ProbabilisticBiasesMeettheBayesianBrain}. A dimension of neuroscience that is lacking in such frameworks is the study of the underlying effects of emotions on cognitive states, or indeed, the function of emotions as heuristics that illicit action. Emotions, for the scope of this text, are particularly interesting when accompanying cognitive miscalculations or biases through secondary inducers (mental objects generated by memory) or through perceiving a primary inducer (external source of stimuli that induces states that are pleasurable or aversive) as mediated by the Amygdala \cite{amygdaladecisionmaking, emotiondecisionmakingorbitofrontralcortex}. Some miscalculations affect, for example, the mental computation of chance, overweighting the certainty of improbable events and underweighting events of moderate probability \cite{Kahneman2000ChoicesVA}. The problem treated herein is made difficult by the confusion that emerges from employing concepts such as "consciousness", "conscious decision", "heuristical thinking", "free will", and so on. For the purposes of this paper, "thinking" or "computation" or any other term related to the act of thinking does not actually mean to impose or suppose any theory regarding "consciousness" and "free will". Instead and unless expressed otherwise, these concepts refer to the computation that occurs alongside or precedes thought - the wetware that enables it.

This work firstly attempts to describe C-CB using a formal logical framework, and secondly to conceptualise a model that predicts the effects of this set of neuronal heuristics. In terms of the latter, a novel statistical model is proposed for simulating decision-making: a Weighted Hidden Markov Model (WHMM). This is the outcome of crossing the Bayesian Neurophysics theory of decision-making as impacted by secondary inducers with cognitive biases and logic. This framework is then a set of isomorphisms between the biology of the brain and its computations and the computation in computers and logic. This proposal is aimed at predictively explaining erroneous human behaviour. Adding to the stuffed bag of 188 confirmation biases is not the end objective of this work. The aim is rather to, firstly, acknowledge a peculiar pattern prevalent in many strategy-seeking and everyday decision-making thought processes, secondly, to formalise the theory using advanced logic and thirdly, to envision a model capable of predicting this aspect of cognition using a novel explainable statistical framework: the Weighted Hidden Markov Model.

\subsection{Self-Fulfilling Prophecy}
\textit{C-Causal Blindness} bears a strong resemblance to the classic notion of a \textit{self-fulfilling prophecy}, while also differing in important ways. Both concepts involve a feedback loop in which an initial belief or strategy inadvertently contributes to the realization of the very outcome one aims to avoid. As Merton famously defined it, a self-fulfilling prophecy is "a false definition of the situation evoking a new behavior which makes the originally false conception come true" \cite{mertonSelfFulProphecy}. In other words, expectations (even if erroneous) can alter behavior such that they confirm the initial expectation. A well-known example is the \textbf{Pygmalion effect} in education: teachers’ unfounded high expectations about certain students can lead them to interact in ways that boost those students’ performance, thereby making the initial false belief come true \cite{rosenthalJacobson1968, oakesExpectations}. More generally, the adage "you see what you seek, you get what you expect" captures the tendency of individuals to interpret information and act in ways that confirm their prior beliefs - a cognitive inclination closely tied to confirmation bias. C-CB shares this self-confirming dynamic: an individual’s policy (action plan) based on a biased expectation brings about the undesired outcome, effectively fulfilling the "prophecy" that if one does \textbf{B} to avoid $\lnot$\textbf{C}, then $\lnot$\textbf{C} will occur. Kurt Gödel's tragic decision to refrain from eating due to fear of being poisoned is a case in point: his preventive action (\textbf{B}) caused the very outcome (\textbf{$\lnot$C}) he feared, exemplifying a self-fulfilling prophecy of the negative kind. However, there are key distinctions. Traditional self-fulfilling prophecies can be positive or negative and often involve social or external expectations (e.g., one person’s expectations about another) \cite{jussimSocialReality}, whereas C-CB refers specifically to an \textbf{intrapersonal cognitive blind spot}: a flawed internal decision-making pattern where one’s own strategy to avoid an outcome leads directly to that outcome. In simpler terms, C-CB is a \textit{self-defeating loop} in one’s reasoning process a special case of a self-fulfilling prophecy engineered by the mind itself in which the very steps taken to prevent an outcome ensure that it happens.

From a cognitive and computational perspective, frameworks such as the \textit{Bayesian brain} model and \textit{predictive processing} theories help illuminate how such paradoxical behavior can arise. The Bayesian brain hypothesis suggests that the mind continually generates predictions (priors) about the world and updates them in a Bayesian manner as new sensory evidence arrives \cite{ProbabilisticBiasesMeettheBayesianBrain}. This view has been successful in explaining many systematic heuristics and biases as byproducts of an otherwise rational probabilistic inference system. Predictive processing, closely related to the Bayesian brain and the free-energy principle, further proposes that the brain not only updates its predictions but also uses them to actively shape perception and action in order to minimize prediction error \cite{clarkSurfingUncertainty, hohwyPredictiveMind}. In other words, the brain is an active agent, inclined to make its predictions come true by filtering interpretations of incoming data and even initiating actions that align with its expectations. Such predictions effectively act as self-fulfilling prophecies, as the brain’s strong expectation triggers behaviors that realize the expected state. While in most cases this mechanism is adaptive (e.g., smoothly guiding routine actions), it can backfire when predictions or priors are distorted by fear or bias. If an agent strongly expects a negative outcome, they may subconsciously select actions that confirm that outcome, since doing so yields less surprise than a disconfirming event. This provides a computational lens on C-CB: a biased predictive system may assign a high prior probability to disaster, and, in a bid to minimize surprise, the cognitive system inadvertently induces the very disaster it aims to avoid. Emotional heuristics, such as fear-driven avoidance, likely amplify these biased predictions, with evidence showing that emotion-processing centers (e.g., the Amygdala) modulate decision-making under uncertainty \cite{amygdaladecisionmaking, emotiondecisionmakingorbitofrontralcortex}.

To formally capture and examine this self-fulfilling bias loop, we turn to computational modeling - in particular, Hidden Markov Models (HMMs). HMMs provide a useful framework for modeling sequential decision processes under uncertainty \cite{russellNorvig, strensBayesianDynamic}. By constructing a \textit{Weighted Hidden Markov Model (WHMM)} that incorporates the agent’s biased decision policy, we can simulate scenarios where an agent’s sequence of actions leads to an unintended outcome. Specifically, the WHMM models how an agent starting from an initial state (e.g., the "current situation") chooses actions based on a policy aimed at achieving a goal state (avoiding hazard \textbf{$\lnot$C}), and how under certain bias weights this policy increases the likelihood of ending in the negative state (\textbf{$\lnot$C}). Such a model allows us to vary parameters (e.g., the strength of the bias or the confidence in the erroneous belief) and observe when the self-defeating outcome prevails. This computational approach aligns with the Bayesian perspective, in which each state transition is interpreted probabilistically, and it makes the abstract concept of C-CB concrete and testable. By observing the model’s behavior, we can identify the conditions under which "the road taken to avoid one’s destiny leads straight to it," and how tweaking the agent’s inferential parameters might break the loop.

Comparing C-CB with the self-fulfilling prophecy highlights that C-CB is likely a \textit{compound cognitive bias} rather than a single isolated error. It encapsulates elements of confirmation bias (seeking evidence for one's fears), availability heuristics (where vivid negative outcomes dominate decision criteria), and anchoring on an initial paranoid premise. This is why we describe C-CB as a bundle of biases: it arises when multiple facets of human cognition - emotional weighting, heuristic simplification, and predictive confirmation - converge to create a vicious cycle. In the following sections, we formalize this peculiar pattern and investigate it through the lens of logic and computation, with the ultimate aim of understanding how such self-sabotaging strategies can be predicted, detected, and potentially averted.

\section{Methods}
An experiment is then proposed to attempt and replicate C-CB in action. This experiment will consist of three phases. The first phase involves presenting the subject with the problem at hand ("Problem" in Fig. 1). In the second phase, the subject is presented with the initial state ("Current State" in Fig. 1) and the goal of the problem ("Goal" in Fig. 1). This problem has a known solution (a proven policy to reach the goal - "B" in Fig. 1). The solution shouldn't be clear to a non-expert, given the previously described assumptions. Together with the correct solution ("B" in Fig. 1), other arbitrary solutions are presented ("C" and "D" in Fig. 1) as well as the opposite of the correct solution ("A" in Fig. 1). Finally, the subject will be presented with a set of policies or plans (the policy: "A", "B", "C", and "D" in Fig. 1) that can be taken to reach the goal. These statements will be one of three types: a statement that will lead the subject directly to the goal (this is known from empirical evidence: "B" in Fig. 1); a statement that results in the opposite of the goal ("A" in Fig. 1) and other arbitrary policies for which there is no real evidence either way ("C" and "B" in Fig. 1).

The subject will read through the set of possible actions, and must immediately decide which one they think will lead them to the goal. The results of the chosen actions are noted. This is repeated for a set of 100 people selected from a random sample of non-experts in the subject of the questions (to be selected by enquiry regarding subject occupations, education, and hobbies). 
A model is then constructed, WHMM, that should predict the effects of C-CB.

The goal here is then to discover if the results of C-CB happen frequently enough that it can be registered as a phenomenon. The following is an example of a problem posed during the experiment:

\begin{figure}[H]
   \centering
   \includegraphics [width=0.70\textwidth] {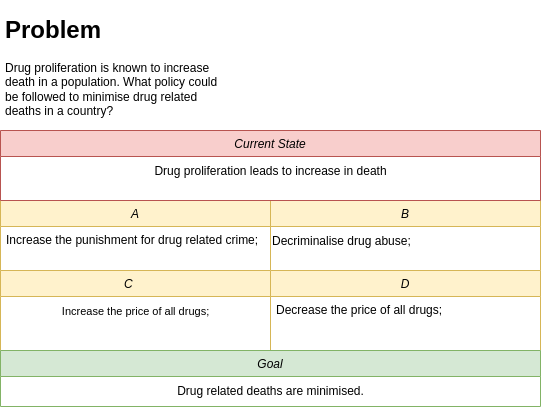}
  \label{fig:label2}
  \caption{Experiment example based on \cite{portugalwarondrugs, usawarondrugs, failurewardrugs}.}
\end{figure}

\subsection{Weighted Hidden Markov Models}\label{WHMM}

A \textbf{Hidden Markov Model (HMM)} extends the concept of a \textbf{Markov chain}, which is a probabilistic model capturing sequences of random variables or states. In a Markov chain, transitions between states follow defined probability distributions, and these states can represent any abstract or concrete entities. Central to the Markov chain is the \textbf{Markov Assumption}: the probability of transitioning to a future state depends only on the current state, not on the sequence of prior states. A standard HMM builds on this by introducing hidden states that generate observable outputs (emissions)\cite{hmm}. It is formally defined by:

\begin{itemize}
    \item \textbf{Q}: A set of \( N \) hidden states.
    \item \textbf{A}: A transition probability matrix, where \( a_{ij} \) denotes the probability of transitioning from state \( i \) to state \( j \), with \( \sum_j a_{ij} = 1 \) for each \( i \).
    \item \textbf{B}: An emission probability matrix, where \( b_{ik} \) represents the probability of observing symbol \( k \) from state \( i \), with \( \sum_k b_{ik} = 1 \) for each \( i \).
    \item \textbf{\(\pi\)}: An initial probability distribution over states, where \( \pi_i \) is the probability that the chain begins in state \( i \), and \( \sum_i \pi_i = 1 \). Some states \( j \) may have \( \pi_j = 0 \), indicating they cannot be initial states.
\end{itemize}

The proposed \textbf{Weighted Hidden Markov Model (WHMM)} adds a factor to the standard HMM by incorporating an additional layer of complexity: \textbf{somatic weights}. These weights represent subjective biases, such as emotional heuristics or cognitive distortions, that influence the agent's perception of state transitions or emissions. WHMM explicitly defines these weights as modifiers of the model's probabilities.

In the WHMM each transition from state \( i \) to state \( j \) is assigned a weight \( w_{ij} \), reflecting the agent's bias toward or against that transition (e.g., overestimating likelihood due to fear or novelty). The agent's perceived transition probabilities are adjusted accordingly:
\begin{equation}
a'_{ij} = \frac{a_{ij} \cdot w_{ij}}{\sum_j a_{ij} \cdot w_{ij}}
\end{equation}
where \( a'_{ij} \) is the biased transition probability, and the denominator ensures normalization (i.e., \( \sum_j a'_{ij} = 1 \)).
Optionally, emission probabilities could also be weighted, with \( b'_{ik} = \frac{b_{ik} \cdot v_{ik}}{\sum_k b_{ik} \cdot v_{ik}} \), where \( v_{ik} \) are emission weights, though this is less emphasized here.

In WHMM, the weights are parameters that distort the agent's internal model of the world, while the hidden states represent the agent's perceived state sequence, and emissions are the observable outcomes or feedback. The true probabilities remain unknown to the agent (hence "hidden"), and the weights encapsulate their subjective distortions.

As an example, an agent might overestimate the probability of a favorable outcome due to biased weights, leading to suboptimal choices. By simulating state sequences with the WHMM, these biases might skew probability estimates and decision outcomes, just as emotional or heuristic factors do in the real world \cite{Kahneman2000ChoicesVA}.

\section{Results}
\subsection{C-CB as a bundle of cognitive biases}
In reasoning and planning, evidence for C-CB is gathered in the willingness to jump to a conclusion or strategy without an explicit analysis of the quality of a plan, as measured by its effectiveness or impact in delivering the desired goal.
This could follow from an underlying weighting of evidence, which starts an erroneous train of thought or causes a misstep-in-logic and in the brain's implicit probability calculation. Regardless of the problem or strategy, activation of brain circuitry is relative to the way that said brain has mapped the relevant reality to the brain's topology in the form of neural networks - circuitry that wires together fires together, but it also maps objects in reality that perceived together\cite{wiretogetherfiretogether}. There are more brain areas involved in these computations than just semantically related circuitry and these are also associated with somatic weighing during decision-making (e.g. the Amygdala). The Amygdala and the Orbitofrontal Cortex are expected, in this context, to negatively influence the decision-making process yielding less-than-optimal decisions as predicted by Damasio et al. \cite{amygdaladecisionmaking, emotiondecisionmakingorbitofrontralcortex}. Two assumptions of C-CB are then, that the subject is not influencing their own thought by formal means (e.g. pen and paper) or otherwise. Secondly, the subject cannot be an expert nor have too much experience in the given context. Not being an expert is then what allows for diverging steps from the goal. This is because, in the case of an expert, their brain will possess some representation of the problem or context that already enables solutions better than that of the random sample. In the case of the non-expert, the guiding processes of thought minus the evidence or data can result in seemingly arbitrary plans to reach the goal. It is the hypothesis of this paper that the results of C-CB are a specific subset of all the ways in which a faulty plan is computed: those policies which result in the inverse of the goal.

\subsection{Examples}
\subsection{I) The (ir)rational student}
Take a state: \textbf{(A)}, where a university student has as their goal (\textbf{C}): finishing their BSc on time.
The student decides to do the following: they take on extra courses to avoid study delays and increase their chances of finishing the degree by stacking backup credits (\textbf{B}).
However, taking extra courses obviously complicates the task by increasing the workload. This increases the likelihood of failing each individual course instead (\textbf{$\lnot$C}).
The student would like to avoid state \textbf{$\lnot$C}, but by following policy \textbf{B} they actually increase the likelihood of ending up in \textbf{$\lnot$C}. Therefore, regarding the overall policy (following \textbf{B} to reach \textbf{C} from \textbf{A}), not doing \textbf{B} might result in a higher chance of reaching \textbf{C}.
That is not to say that there isn't an optimal policy, \textbf{$B^{*}$}, for which the probability of reaching \textbf{C} is not maximised, but rather to notice the peculiar symmetrical nature of the whole ordeal: that one would come up with a strategy that indeed increases the likelihood of the inverse of the goal.

\subsection{II) The war on drugs}

\subsubsection{The initially perceived causality chain (\textit{relaxed} \textbf{B})}
\begin{enumerate}
\item \textbf{A} = Drug proliferation leads to increase in death;
\item \textbf{B} = Heavily punish drug related crime;
\item \textbf{C} = Drug related deaths are minimised.
\end{enumerate}

\subsubsection{The causality chain as an outcome of \textit{relaxed} \textbf{B} (inverted or negation of \textbf{C}) (after testing the hypothesis or implementing the policy)}
 \begin{enumerate}
\item \textbf{A} = Drug proliferation leads to increase in death;
\item \textbf{B} = Heavily punish drug related crime;
\item \textbf{$\lnot $C} = Increase in drug related deaths \cite{urbanrural}.
\end{enumerate}

\subsubsection{The causality chain with \textit{corrected} \textbf{B} }
 \begin{enumerate}
\item \textbf{A} = Drug proliferation leads to increase in death;
\item \textbf{$\lnot $B} = Decriminalise drug abuse;
\item \textbf{C} = Decrease in drug related deaths \cite{portugalwarondrugs}.
\end{enumerate}

In this case, the US's infamous strategy just happens to be the exact opposite of Portugal's which maps well to C-CB as an example. While the US's strategy has failed and indeed might have led to increases in death \cite{failurewardrugs}: $\lnot$ C, Portugal's, having achieved \textbf{C}, has been a success, being hailed by worldwide health organisations as a gold-standard policy \cite{portugalwarondrugs}. This is a good example of C-CB as long as one understands that countries don't make legislation - people do. And it's people who suffer from the sort of cognitive bugs that would result in this \textit{cognitively relaxed policy} with an erroneous \textit{causal chain}.
The important point here is rather to explicitly procure examples so as to demonstrate that not only is \textbf{B} erroneously deduced to cause \textbf{C}. For the US, the policy resulted from the belief that drug laws and criminal punishments were too light, but applying \textbf{B} in fact caused \textbf{$\lnot $C}: the opposite of the goal state, leading to contradiction, which is solved in Portugal's case by using \textbf{$\lnot$ B} or loosening drug legislation and criminal punishment yielding \textbf{C}. Noticing that indeed the opposite of what was intended happened, and knowing that C-CB occurs as the result of a bundle of cognitive biases, yields a foundation for claiming that such strategies are then bad strategies and the methods that accomplished them (attested by the infamous background for the current war on drugs movement of the USA \cite{usawarondrugs}) were biased and even irrational. In other words, C-CB is a sufficient base on which to disqualify an argument.

\subsection{Logical dimension}
Using possible-worlds theory: there is a world \textbf{B} that is accessible from world \textbf{A} (the current state made up of a set of facts). Take now a third world - world \textbf{C}, only accessible from \textbf{B}. Another set of worlds exists, where instead, \textbf{C} is not accessible from \textbf{B} but \textbf{$\lnot $C} is. Finally, a third set exists,  where \textbf{$\lnot $B} is the world accessible from \textbf{A} that could lead to \textbf{C} (Figure 1).

\begin{figure}[h!]
   \centering
   \includegraphics [width=0.50\textwidth] {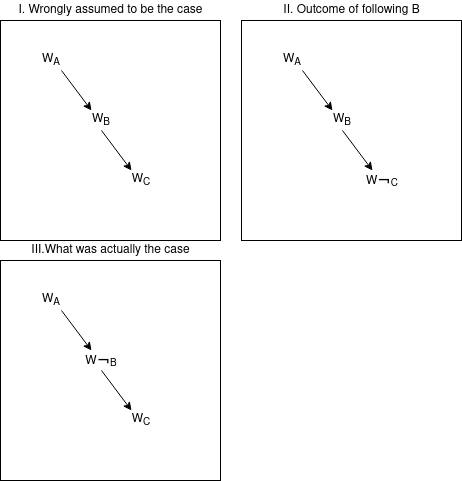}
  \label{fig:label3}
  \caption{Three possible worlds scenario}
\end{figure}

What is being referred to as \textit{C-CB} in this paper is a cognitive mistake that supports, through feelings of certainty or familiarity, the causal chain between B and C, that is, the erroneous prediction that following policy B will yield C - what is referred here as a (cognitively) relaxed causal chain or relaxed picture of the universe. Cognitively relaxed because no guided or active cognitive work is done apart from an initial spontaneous or innate computation that yields the policy. The three sets of worlds are used to illustrate three potential universes:
\begin{itemize}
 \item The relaxed mental universe 1 (or the hypothetical), where the belief error occurs ($B \rightarrow C$);
 \item The real universe 2, where the hypothesis is empirically tested ($B \rightarrow \lnot  C$);
 \item The corrected mental universe 3 (or the hypothetical that is in line with the empirical universe), where the belief error is either substituted by some B with less likelihood of being an error, more likely to result in \textbf{C} or by the optimal causal chain that matches universe 2 - the inverted chain ($\lnot  B \rightarrow C$). The latter may only become obvious in retrospect, after empirically testing the policy.
\end{itemize}

\par
This relaxed picture of the universe is an inversion of the second causality chain in strategy-based or goal-oriented thinking and planning. This results in predicting that $B \rightarrow C$, when in fact, $(\lnot  B \rightarrow C) \land (B \rightarrow \lnot  C)$ is the case. \textbf{B} emerges relatively easily in terms of cognitive work and it is an accessible state from the current state \textbf{A} - in fact, it is the topology of brain networks that yields, potentially, any accessible state that is even loosely related to \textbf{A} and \textbf{C} \cite{wiretogetherfiretogether}. \textbf{C} is erroneously deduced (or its likelihood grossly overestimated) as a direct effect of \textbf{B} when there is minimal to no evidence to substantiate this belief. This objectively wrong causal chain idealised as correct that exists between \textbf{B} and \textbf{C} is the outcome of known mental fallacies like the \textit{appeal to novelty} fallacy or the \textit{ambiguity effect}. The following is another example that hopefully helps to ground further the abstract situation described above.

\subsection{Modal Logic Expansion}


Modal Logic is often used to analyse reasoning involving necessary conditions as well as possible conditions. C-CB is reducible to a case where a possible outcome is being confused for a necessary outcome when following one of many possible policies to reach a certain goal. This general form can be represented using Modal Logic as follows:\\

Relaxed Form)\footnote{Where $\Box$ is the modal operator for "it is necessary that".}

\begin{center}
\begin{math}
\Box ( B \rightarrow C)
\end{math}
\end{center}
The subject assumes some policy \textbf{B} (through arbitrary reasoning - i.e. equivalent to picking a random plan from a set of plans that involve the themes of the problem) to be ideal - this is simply because it is not part of the cognitive toolset to experiment with idealised policies in the head before deploying them in the world (unfortunately). In other words, there is no experimental phase before \textbf{B} is taken to be optimal (chosen) as a plan to reach \textbf{C} - the cognitively relaxed causal chain. It is also assumed that by executing policy \textbf{B}, \textbf{C} must then be necessarily the case without any evidence whatsoever apart from a heuristically motivated feeling.
What can be verified in reality is quite fuzzier and less straightforward than the idealised causal chain:\\

What Is Probably The Case)\footnote{Where the operator $\Diamond$ signifies ‘possibly'. }
\begin{center}
\begin{math}
\lnot  \Box ( B \rightarrow C) \land \Diamond (  B \rightarrow C) \land \Diamond(B \rightarrow \lnot C)
\end{math}
\end{center}


Applying this to example II), it is not necessarily the case that the increase in legislation and increased severity of drug-related crime punishments (\textbf{B}) will lead to a decrease in drug-related deaths (\textbf{C}). In fact, the opposite occurs: following \textbf{B} results in \textbf{$\lnot $C} - this is C-CB. But ultimately it can be the case that the cognitively relaxed policy does work. In other words, it is possibly the case that following \textbf{B} does result in \textbf{C}, but this might as well be the result of chance and not of correct problem-solving etiquette - this is not an example of C-CB but a superset of C-CB: the set of all plans that when executed will go wrong. Regardless of achieving a correct outcome, it is the lack of explicit probabilistic reasoning and not having empirical evidence as a prerequisite for belief at the root of thought as well as a natural predisposition to lean on heuristical thinking which is yielding a cognitively relaxed policy. It is not strange that this would be the case. In an evolutionary context, selection, which applies to all dimensions of the organism, prioritises the minimisation of energy expenditure. Thus, the brain has a preference for firing circuitry that does not need much or any attention at all \cite{attention1, attention2}. The conscious mind is then a product of this default: loosely related statements that are contextually involved in arbitrary M-dimensions that vary (are relative) from person to person. The guiding processes of which are not dissimilar to a blind pilot at the helm of a kart, driving full speed ahead, bouncing from heuristical wall to heuristical wall like a pinball machine while hoping to cross the finish line. This is the true cause of C-CB and other heuristics: nature. It is due to the patchy work of nature and the absence of an explicit programmer that brains are essentially dumb while being the most complex computational object known to us.


\section{Discussion}
The exploration of C-Causal Blindness (C-CB) as a cognitive phenomenon opens several avenues for future investigation, both empirical and theoretical, to solidify its standing as a distinct bias bundle and refine its predictive modeling using WHMMs or otherwise. A primary direction for future work lies in conducting the experiment outlined in the paper. This involves recruiting a large sample of non-expert participants and presenting them with a problem, an initial state, a goal, and a set of policies (e.g., A, B, C, D), where B is the correct policy, A leads to the opposite outcome ($\not C$), and C and D are arbitrary. By analyzing the frequency of policy selections, particularly the tendency to choose A over B it could be assessed whether participants systematically favor policies that invert their intended goals, thereby validating C-CB as a measurable cognitive misstep.

Beyond this foundational experiment, future research could explore the usefulness of WHMM by implementing it computationally and fitting it to the collected experimental data. The WHMM, with its somatic weights reflecting emotional or heuristic influences, offers a novel approach to simulating decision-making under C-CB. By adjusting parameters, such as the strength of biases like the appeal to novelty or ambiguity effect, researchers can test the model’s ability to predict participant choices accurately. This computational step would bridge the theoretical framework with observable behavior, allowing for iterative refinement of the model. For instance, if the WHMM consistently predicts a higher likelihood of selecting policy A when emotional weights are amplified, this could underscore the role of affective heuristics in driving C-CB, aligning with the paper’s hypothesis about the amygdala and orbitofrontal cortex.

Another promising direction involves investigating the neural underpinnings of C-CB. Given the speculated involvement of emotion-processing regions like the amygdala and decision-making areas like the orbitofrontal cortex, future studies could employ neuroimaging techniques such as functional Magnetic Resonance Imaging (fMRI) to identify brain activity patterns associated with C-CB choices. For example, heightened amygdala activation during the selection of policy A (leading to $\not C$) might indicate fear-driven avoidance amplifying the bias. Comparing these neural signatures across experts and non-experts could further elucidate whether expertise mitigates C-CB, as hypothesized. If experts exhibit reduced activation in these regions or greater engagement of prefrontal areas linked to analytical reasoning, this would suggest that training or experience rewires the cognitive processes prone to C-CB.

Finally, the evolutionary and real-time detection aspects of C-CB warrant deeper exploration. From an evolutionary perspective, if C-CB stems from a preference for quick, low-effort decisions, as might have been advantageous in ancestral environments, future work could examine its presence in other species or across human populations with varying ecological pressures. Simultaneously, developing methods to detect C-CB in real-time, perhaps through behavioral cues (e.g., hesitation patterns) or computational analysis of decision sequences, could transform it from a retrospective observation into a proactive diagnostic tool.

Regarding the definition of C-CB, four points are taken to be the most relevant in accessing its weaknesses. These are explored below.

\subsection{1) C-CB might only be confirmed in restrospect}
Had the (ir)rational student taken more courses and passed all of these (as it is bound to be the case for many students) would this still be a case of C-CB or not? If a policy succeeds, does this not mean that the erroneous thought patterns were in fact correct? And a policy that would've been seen as a bad policy (a policy that decreases the chances of their goal when compared to that of the random policy) must be reinterpreted as a good policy, maybe the best policy? More importantly, if C-CB is only to occur in retrospection, then is it a theory at all or just a mildly interesting remark? It might be the case that the policy only worked by chance or "luck" and that indeed it was a terrible policy to begin with. But if C-CB is only concerned with policies that lead to the opposite of the goal, how do these erroneous plans (resulting in the goal or not) relate to C-CB? And how can C-CB be seen except in retrospection?

\subsection{2) Is C-CB really happening in the brain or does it only happen in philosophical discussion?}
Unfortunately, the brain has not been deciphered. Computation is a useful term but to think of what neurons are doing as a true isomorphism to calculations that happen inside the computer might very well be an erroneous misstep. The idea that computer computations are exactly descriptive of reality permeates today's scientific dialogue and research itself, with plenty of theories assuming this analogous relation in many different fields, implicitly or explicitly. This is perhaps most common in physics \cite{Susskind1994TheWA, wolfram1985undecidability}, but also permeates cognitive science \cite{shagrir2006we}, and biology \cite{link2002bugs} to cite just a few. All of these works deploy the type of language which assumes the isomorphism between computation and some real phenomena to hold. Whether true or not in the case of these studies, it is not clear, by any measure, if the assumption holds in the case of the brain and C-CB. This is not to say that the brain is not deterministic, but rather that the mapping between our computational theories and the brain is filled with unknowns. In this sense, it might be the case that C-CB along with other (or all) cognitive biases are just abstract interpretations loosely connected or not to the mechanisms present in the brain. As far as research goes, all cognitive biases and fallacies are simply high-level descriptors of something that is happening involving cognition. There is no explained connection between these theories and the underlying brain mechanisms. And still, the theories hold predictive, replicable value.
\subsection{3) What is the evolutionary argument for C-CB?}
The brain is working with limited resources both at the input level (e.g. information gathered from the senses or through the recursive use of memory like the phonological loop) and in its generative capacity for thought. Language is used to enhance and aid in determining the quality of thought - this comes with increased attentional costs \cite{attention1, attention2}. To add on top of this, in planning, one is faced with a difficult degree of complexity (e.g. questioning what is knowable, grasping complexity issues, associating knowns and unknowns, and the unequivocal difficulty in generating appropriate probabilistic spaces or distributions to associated or causally chained events). Having limited resources yields sub-optimal solutions to computational problems in the brain - this has been widely demonstrated by Daniel Kahneman and many others in Psychology and Cognitive Science \cite{daniel}. The brain's main objective, apart from central control of the body, seems to be the modelling and subsequent prediction of immediate reality so as to allow for goal formation, hypothesising the set of possible solutions to reach said goal, the search for these, and their execution. Most of this seems to happen unconsciously and automatically. C-CB appears to be a combination of priming bias, where the thinker's brain automatically and due to brain mechanics (not due to the thinker's will) is limiting their own set of possible solutions down to one - that which is selected is erroneously taken to be the optimal one and followed blindly. That is, the bound or limit to the set of possible solutions is a momentary brain state composed of information that is accessible to the conscious brain. Confirmation bias then limits the set of possible answers by acting as a funnel that results in the erroneous scenario where a strategy, \textbf{B}, enables a goal, \textbf{C}.

\subsection{4) Why is C-CB a problem and can it be solved?}
It should be immediately apparent how useful this is in a state of nature but how detrimental it might be in a complex modern world. Most people do not have the explicitly learned rational mechanisms (i.e. scientific methodology) or critical thinking tools in place to ensure corrective thinking patterns after planning or to address the issue of the likelihood of its certainty. In other words, people don't normally stop and think about arising thoughts and feelings - there is a lack of meta-analysis. It seems to be the case that selection has evolved the brain to produce useful thinking as measured by quick deliberation and subsequent action under the survival of the fittest rule. But in planning, error correction techniques and testing are obviously favoured. To engage with these higher conceptual problems, meta-heuristical rules must be deployed after thinking and planning. When this does not happen, the thinker makes leaps in judgement, having ignored the automatic origin of the initial plan. They assume that a policy or plan \textbf{B} leads to \textbf{C} from the current state \textbf{A} without resorting to testing or evidence of any kind except for a feeling of familiarity with the concepts or similarity between them. It is the case that people can be taught how to engage in nuanced, rational, and empirically based thinking, else no field of science would have gone very far. This often requires self-consciousness and, ironically enough, learned heuristics, that enable stages of thinking that work as in-between checks and balances occurring during thinking. Writing while thinking enables this type of corrective behaviour through rewrites and rereads. Finally, admitting that biases are causes of thought - that they precede thought and indeed enable it - does not mean that an erroneous conclusion cannot be deliberated about, reformulated, or dropped completely. One only has to doubt their solution to a mathematical problem to set in motion the sort of correcting behaviours that will increase the likelihood of this not being so (go over each step, redo the step, use a calculator, check the solution sheet, etc). It is also the case that, having been wrong on paper, there is always a possibility of erasing the mistake or starting anew. Thoughts are ephemeral and at a minimal cost, one can just think again. Learning the basic notions that enable these corrective patterns of thinking to arise during planning, namely critical thinking and more relevantly, the scientific method, is an obvious advantageous skill. Being aware of what one does not know or is unsure about is important.
\newpage
\section{Conclusion}
Regardless of corrective pathways for this hereditary disease we all might suffer from, C-CB as a concept, if it exists, it does so in the frontier of different sciences, where further research, namely in the fields of Cognitive Computational Neuroscience and AI, is necessary. To better understand human cognitive faults in a predictive fashion, new experiments that enable modelling internal computations, as well as the proper isomorphism between the cognitive dimension, computation, and its logical representations (which are easily rectifiable) are important. Together with evidence from experiments, this computational-logical framework may yield easily interpretable and falsifiable theories regarding the sentimental life of the black box that is the brain.
Here, the mind has been taken to be the computational dimension of the brain, which does not necessarily follow logical rules or proper probability calculations. This results in what has been addressed as relaxed causality chains. Animals, including humans, are encapsulated agents with information being mediated through extremely sensitive sensors, forever without access to the source of the information. The brain and its computational dimension ultimately can and indeed tend to deal with ideals so as to avoid getting stuck and being incapable of making decisions in the face of so many unknowns and continuous sources of input. Most animals are required to make decisions regardless of limited information processing, and they have to make these quickly in scenarios where higher-order cognitive dimensions do not exist or are extremely limited (e.g. no access to complex language with which to abstract the problem, facilitating its solution). Utilizing different analytical techniques (First Order Logic, Modal Logic, and Graph Theory - HMM) to map the cognitive dimension where human reasoning occurs, a framework is proposed with which to check a flawed logical model that might naturally emerge when people think: C-CB. In sum, aspects of the content of the mind have been grounded on logic and therein flawed models that are erroneously assumed to be true may be disproved using WHMM. That is, causality between a current state, a strategy, and a desirable state is simply assumed to exist without any actual logical basis or data to verify this existence. In experiments involving C-CB thought patterns, it can be quickly verified that in order to prevent the erroneous actions induced by C-CB one only needs to know the footprint of a C-CB-enabled thought pattern to correct it: that an initial plan involving $A \rightarrow B \rightarrow C$ shouldn't be assumed to be correct immediately and in fact may lead to the opposite of the goal, \textbf{$\lnot C$}. I must end by disagreeing with Professor Kahneman, who has been quoted as saying that even after his long tenure in pointing out the flaws in human cognition, he himself had failed to activelyt curb any of them in his daily life. I think we must be more pragmatic - once one begins to doubt their own thoughts, stepping onto the meta-analyses, verifying that \textbf{B} lacks evidence is straightforward. To think and evaluate the proposed solution and critically deduce the possible outcomes one does not need any knowledge of probability theory, logic, graph theory, and causal calculus. But these and access to a computer is always a plus.

\textcolor{white}{.}

\newpage
\printbibliography
\newpage



\end{document}